\documentclass[11pt]{article}

\usepackage[margin=1in]{geometry}
\usepackage[T1]{fontenc}
\usepackage{lmodern}
\usepackage{microtype}
\usepackage{amsmath}
\usepackage{booktabs}
\usepackage{caption}
\usepackage{float}
\usepackage{graphicx}
\usepackage{multirow}
\usepackage{natbib}
\usepackage{xcolor}
\usepackage{hyperref}
\usepackage{url}

\definecolor{linkblue}{HTML}{245A8D}
\hypersetup{
  colorlinks=true,
  linkcolor=linkblue,
  citecolor=linkblue,
  urlcolor=linkblue,
  pdftitle={Scale-to-Dialogue: Low-Burden Elicitation of Daily Premenstrual Symptom Ratings with Small Language Models}
}

\newcommand{\doi}[1]{\href{https://doi.org/#1}{doi:#1}.\\[0.35em]}

\title{Scale-to-Dialogue: Low-Burden Elicitation of Daily Premenstrual Symptom Ratings with Small Language Models}
\author{Yifan Wang\\The Chinese University of Hong Kong, Shenzhen (CUHK-Shenzhen)\\\texttt{224050081@link.cuhk.edu.cn}}
\date{}

\begin{document}
\maketitle

\begin{abstract}
Prospective daily symptom tracking is central to premenstrual health assessment, but repeated ordinal forms impose substantial response burden. We formulate conversational administration as an ordinal label-recovery problem: the system actively elicits a small set of symptom clusters and maps each response to the original severity labels. We used 3,320 complete participant-days from the mcPHASES dataset, covering cramps, mood swing, fatigue, sleep issues, stress, and bloating on a six-level scale. Six participants were reserved for development and 36 for a frozen evaluation comprising 360 participant-days and 2,160 item labels. A ModernBERT evidence gate detected whether a symptom was expressed, and Qwen2.5-1.5B-Instruct produced deterministic structured severity scores. Fixed six-item questioning achieved a quadratic weighted kappa of 0.976, whereas three joint symptom-cluster questions achieved 0.913, 97.45\% agreement within one severity level, and 80.94\% recall for moderate-or-higher symptoms while reducing questions by 50\%. Open-first adaptive policies required 3.92--5.98 questions and produced lower agreement than the corresponding fixed policies. Participant-cluster bootstrap analysis estimated a kappa difference of $-0.062$ (95\% CI $-0.076$ to $-0.048$) between the three-cluster and six-item strategies. Active cluster-level elicitation provides a direct, local-model route from natural conversation to reusable daily symptom labels.
\end{abstract}

\noindent\textbf{Keywords:} premenstrual symptoms; conversational assessment; patient-reported outcomes; small language models; ordinal agreement

\begin{figure}[t]
  \centering
  \includegraphics[width=\textwidth]{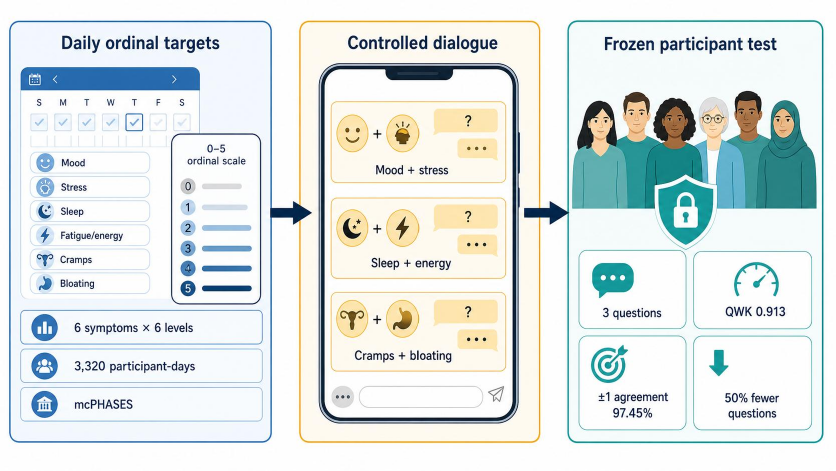}
  \caption{Scale-to-dialogue study overview. Six daily symptom targets spanning six ordinal severity levels were compressed into three controlled symptom-cluster questions and evaluated on a frozen participant test. The three-question policy reduced daily question burden by 50\% while retaining high ordinal agreement.}
  \label{fig:overview}
\end{figure}

\section{Introduction}

Premenstrual symptom assessment depends on repeated measurements. The Daily Record of Severity of Problems (DRSP) operationalizes daily symptom and impairment ratings, and prospective scoring systems such as the Carolina Premenstrual Assessment Scoring System derive cycle-level patterns from those daily observations \citep{endicott2006drsp,eisenlohr2017cpass}. Daily ratings capture within-person changes that retrospective reports miss, but intensive diaries also create a recurring data-entry task \citep{bosman2016daily}. Menstrual tracking research similarly shows that people combine formal records with awareness of mood, pain, and bodily changes, suggesting that symptom capture should fit ordinary self-expression rather than require a separate reporting ritual \citep{epstein2017tracking}.

Computerized adaptive testing reduces patient-reported outcome burden by selecting fewer items while preserving score information \citep{harrison2023emcat}. Conversational systems offer another route: a response can contain evidence for more than one structured field. Prior work has used dialogue systems for functional assessment \citep{sheng2023dialogue}, adapted PHQ-9 and GAD-7 questions with ChatGPT-4 \citep{liu2025agreement}, and demonstrated strong concordance between conversational and form-based PHQ-9 administration \citep{guo2026hopebot}. Recent design work has also proposed topic-level dialogue that fills multiple questionnaire fields from one exchange \citep{navarro2026questionnaire}. These studies motivate conversational assessment, but they do not establish how much item burden can be removed while preserving daily premenstrual symptom labels with locally deployable models.

We therefore treat scale-to-dialogue conversion as a controlled measurement task (Figure~\ref{fig:overview}). Each dialogue policy must cover six symptom domains, output one ordinal label per domain, and expose a measurable trade-off between question count and label agreement. We compare fixed item-level elicitation, fixed symptom-cluster elicitation, open-first adaptive follow-up, and an unconstrained opening-only condition. The models are small enough for local deployment: a bidirectional encoder performs evidence gating and a 1.5-billion-parameter instruction model performs structured ordinal scoring.

\section{Methods}

\subsection{Dataset and participant split}

We used mcPHASES, a longitudinal dataset containing wearable, hormonal, metabolic, and self-reported menstrual health data from 42 Canadian adults who menstruate \citep{lin2026mcphases,lin2025physionet}. The original study collected daily symptoms on a six-point scale and was approved by the University of Toronto Research Ethics Board (Protocol 41568). Participants provided written consent for study participation and public release of de-identified data.

Our processed dataset contained 5,659 participant-days. Complete observations for cramps, mood swing, fatigue, sleep issues, stress, and bloating were available on 3,320 days. Labels were encoded as 0 (not at all), 1 (very low), 2 (low), 3 (moderate), 4 (high), and 5 (very high). The six participants in the source validation split, contributing 453 complete days, formed the development partition. The remaining 36 participants were locked before the final policy evaluation. We sampled 10 days per locked participant, stratified by the number of symptoms rated at least 3, yielding 360 days and 2,160 item labels.

\subsection{Label-conditioned language corpus}

We wrote one direct, one colloquial, and one indirect Chinese seed utterance for every symptom--severity pair, producing 108 seeds. Qwen2.5-1.5B-Instruct \citep{qwen2025technical} rewrote each seed in four styles: chatty, hesitant, event narrative, and brief. Generation preserved the source symptom, severity, negation, and functional impact, and was prohibited from emitting scores or diagnostic terms. Of 432 generations, 364 passed parsing, length, duplication, and score-leakage checks.

Two independent scorers, Qwen2.5-0.5B-Instruct and Qwen2.5-1.5B-Instruct, then scored each generated utterance without access to its label. An utterance was retained when both predictions were within one level of the source label and the two predictions differed by no more than one level. This retained 298 utterances; 66 were placed in a review queue. The retained corpus was used only to develop the language components and was not added to the locked participant labels.

\subsection{Evidence gate}

Each retained utterance was paired with all six target symptoms. The source symptom formed a positive pair and the other five symptoms formed negative pairs. We fine-tuned ModernBERT-base \citep{warner2024modernbert} as a binary sequence classifier using chatty and brief styles for training, hesitant utterances for threshold selection, and event narratives for testing. Training used six epochs, batch size 32, maximum sequence length 128, AdamW with learning rate $2\times10^{-5}$, gradient clipping at 1.0, and class weights of 1:5 for negative and positive pairs. At each epoch, the operating threshold maximized validation recall subject to a false-positive rate no greater than 0.05. We repeated training with seeds 20260809, 20260810, and 20260811.

The locked dialogue evaluation used the model from seed 20260809 and its frozen threshold of 0.435. The gate received a target symptom and the user's utterance. Explicit absence counted as evidence because it supports a zero rating; silence about the target did not.

\subsection{Ordinal severity scorer}

We used Qwen2.5-1.5B-Instruct as the final severity scorer. The system prompt defined the six ordinal anchors through symptom intensity and functional interference. Decoding was deterministic. A single-item response produced one JSON score. A cluster response used a joint prompt that named both target symptoms and required two independent JSON scores; the prompt explicitly prohibited transferring one symptom's severity to the other.

The 1.5B scorer was selected on 120 development days. Relative to the 0.5B model, it increased QWK from 0.713 to 0.977 for fixed six-item questioning and from 0.844 to 0.924 for fixed three-cluster questioning. Prompts, cluster composition, sampling, and evaluation code were then frozen.

\subsection{Dialogue policies}

We evaluated five policies. The fixed six-item policy asked one targeted question for each symptom. The fixed three-cluster policy asked one question for mood swing and stress, one for fatigue and sleep issues, and one for cramps and bloating. Each answer produced two separate scores. The open-plus-item policy began with a free narrative and then asked one question for every item not resolved by the evidence gate. The open-plus-cluster policy used the same opening but covered unresolved items through the three symptom clusters. The opening-only policy scored only symptoms detected in the initial narrative.

For offline evaluation, each participant-day generated an initial narrative from the highest-severity symptoms: up to two symptoms rated at least 3 were disclosed; otherwise the highest nonzero symptom was disclosed, and all-zero days produced a neutral statement. Follow-up responses were deterministic label-conditioned utterances. This design held symptom content constant while isolating the effects of evidence gating, joint scoring, and question policy.

\subsection{Outcomes and statistical analysis}

The primary outcome was quadratic weighted kappa (QWK) between predicted and source labels \citep{cohen1968weighted}. Secondary outcomes were mean absolute error (MAE), exact agreement, agreement within one severity level, score coverage, and recall for moderate-or-higher symptoms (labels 3--5). Question count included the initial opening when present.

We compared each clustered policy with fixed six-item questioning using 2,000 paired bootstrap replicates clustered by participant \citep{efron1993bootstrap}. Each replicate sampled participants with replacement and retained all selected participant-days and symptom items. We report percentile 95\% confidence intervals. A QWK difference of $-0.05$ was used as the non-inferiority margin for the compact policy. Experiments ran in Python 3.11 with PyTorch 2.8.0, Transformers 5.3.0, NumPy 2.3.5, pandas 2.3.3, and scikit-learn 1.8.0 on one NVIDIA GeForce RTX 5090 D GPU.

\section{Results}

\subsection{Evidence detection}

Training the evidence gate on direct hand-written templates did not transfer to indirect language: test recall was 0.111 at a false-positive rate of 0.017. Replacing that training set with the consensus-filtered generated corpus increased event-narrative recall to $0.749\pm0.042$, precision to $0.850\pm0.059$, and F1 to $0.794\pm0.005$ across three seeds; the false-positive rate was $0.027\pm0.014$.

On the 360 locked initial narratives, 511 symptoms were disclosed. The frozen gate produced 315 true positives, 52 false positives, 196 false negatives, and 1,597 true negatives, corresponding to 0.616 recall, 0.858 precision, and a 0.032 false-positive rate. The 52 false-positive gates also produced 52 unsupported initial severity scores.

\begin{figure}[htbp]
  \centering
  \includegraphics[width=\textwidth]{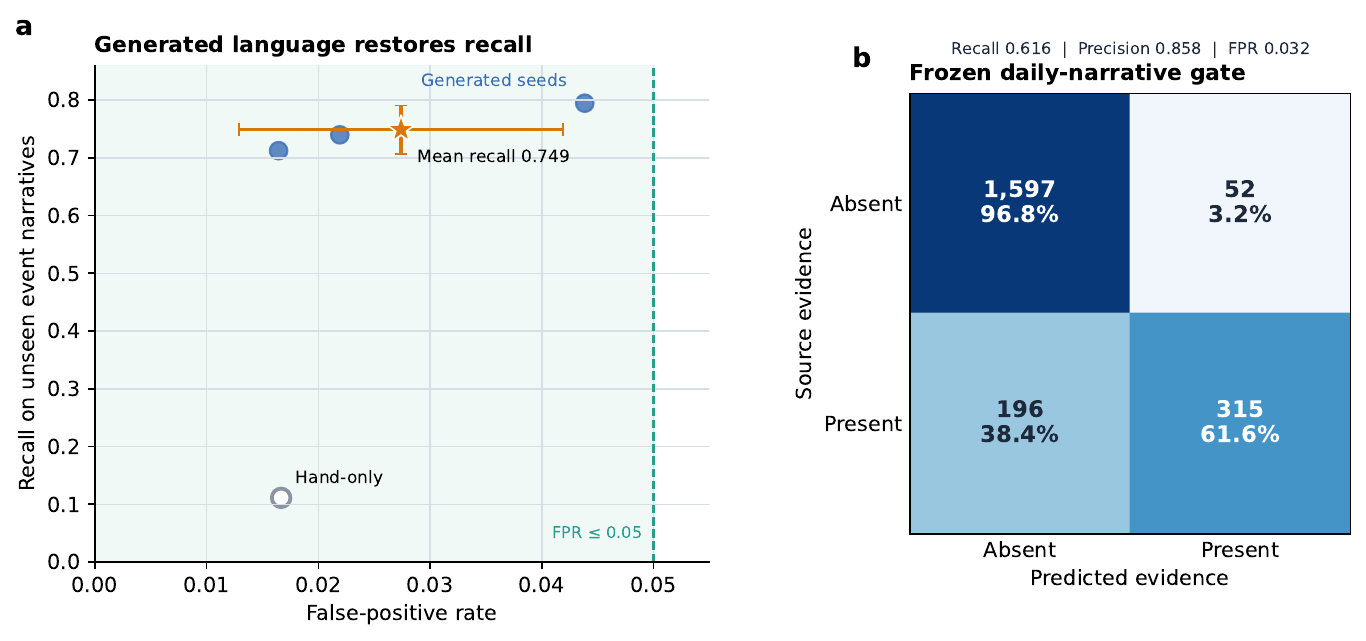}
  \caption{Evidence-gate generalization and frozen performance. \textbf{a}, recall versus false-positive rate on unseen event narratives for hand-template training and three generated-corpus seeds; the star and error bars show the generated-corpus mean and standard deviation. \textbf{b}, row-normalized confusion matrix and operating metrics for the frozen seed-20260809 gate on 360 locked daily narratives.}
  \label{fig:gate}
\end{figure}

\subsection{Dialogue policy comparison}

Fixed six-item questioning produced the highest label agreement (QWK 0.976; MAE 0.087; exact agreement 91.30\%). The fixed three-cluster policy halved the question count, with QWK 0.913, MAE 0.251, exact agreement 77.45\%, and within-one agreement 97.45\% (Table~\ref{tab:policy}). Its moderate-or-higher recall was 80.94\%, compared with 76.87\% for fixed six-item questioning.

Open-first policies did not improve the efficiency--agreement trade-off. Open-plus-item questioning required 5.98 questions and reached QWK 0.943. Open-plus-cluster questioning required 3.92 questions and reached QWK 0.886. The opening-only policy covered 16.99\% of labels; QWK among covered labels was 0.492.

\begin{table}[t]
\centering
\caption{Locked participant evaluation of dialogue policies.}
\label{tab:policy}
\small
\begin{tabular}{lrrrrrrr}
\toprule
Policy & Questions & Coverage & QWK & MAE & Exact & Within 1 & Recall $\geq3$ \\
\midrule
Fixed six items & 6.00 & 100.00\% & 0.976 & 0.087 & 91.30\% & 100.00\% & 76.87\% \\
Fixed three clusters & 3.00 & 100.00\% & 0.913 & 0.251 & 77.45\% & 97.45\% & 80.94\% \\
Open + item follow-up & 5.98 & 100.00\% & 0.943 & 0.145 & 87.82\% & 98.61\% & 83.06\% \\
Open + cluster follow-up & 3.92 & 100.00\% & 0.886 & 0.296 & 75.09\% & 96.25\% & 87.13\% \\
Opening only & 1.00 & 16.99\% & 0.492 & 0.575 & 56.13\% & 91.83\% & 96.03\% \\
\bottomrule
\end{tabular}

\vspace{2pt}
\parbox{0.96\textwidth}{\footnotesize QWK, MAE, exact agreement, within-one agreement, and recall for the opening-only policy were calculated only on covered labels.}
\end{table}

\begin{figure}[htbp]
  \centering
  \includegraphics[width=\textwidth]{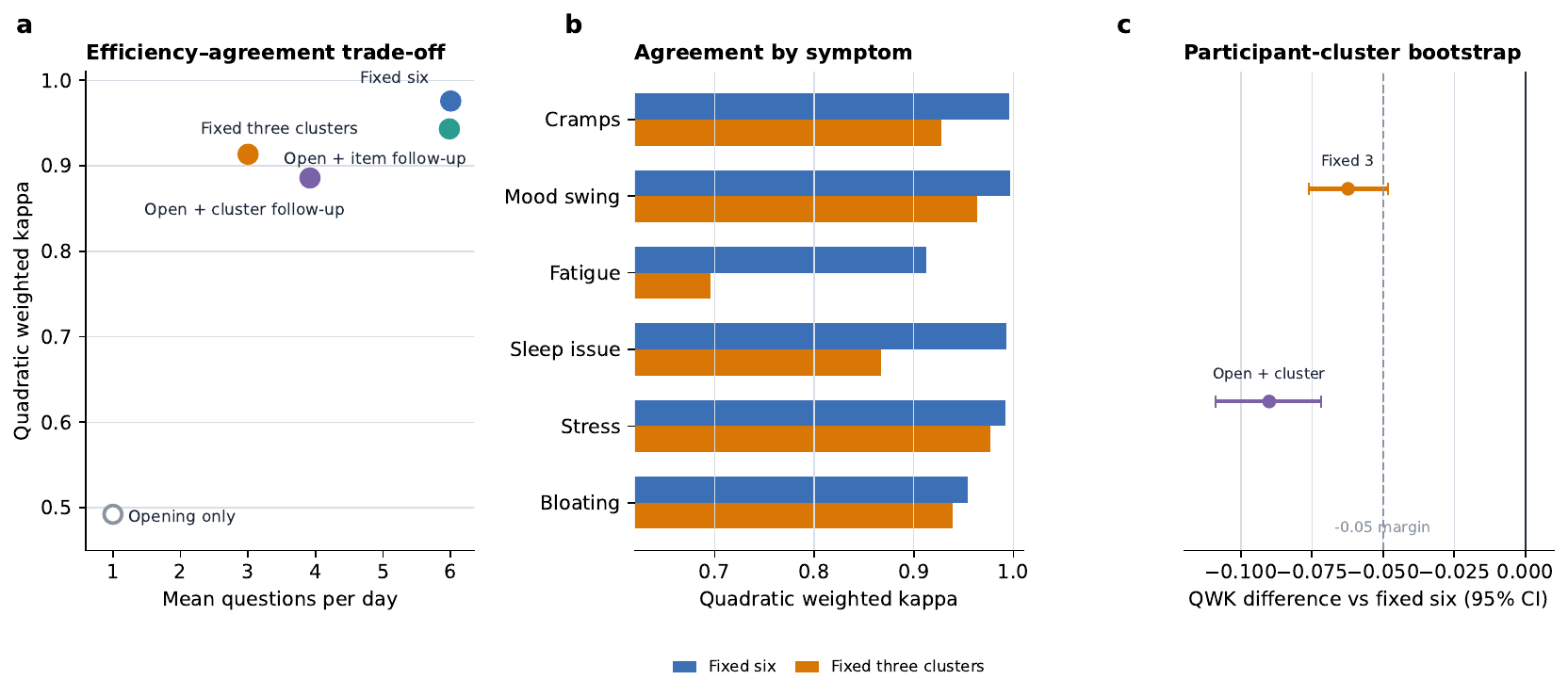}
  \caption{Question efficiency and ordinal agreement on the locked participant split. \textbf{a}, Mean daily questions versus QWK for five dialogue policies; the hollow opening-only marker denotes 16.99\% coverage. \textbf{b}, symptom-level QWK for fixed six-item and fixed three-cluster questioning. \textbf{c}, participant-cluster bootstrap differences in QWK relative to fixed six-item questioning; points show estimates and bars show percentile 95\% confidence intervals. The dashed line marks the $-0.05$ non-inferiority margin.}
  \label{fig:tradeoff}
\end{figure}

The participant-cluster bootstrap estimated a QWK difference of $-0.0623$ (95\% CI $-0.0761$ to $-0.0483$, two-sided $p=0.001$) for fixed three clusters versus fixed six items. The confidence interval crossed the $-0.05$ non-inferiority margin. Open-plus-cluster follow-up had a QWK difference of $-0.0900$ (95\% CI $-0.1089$ to $-0.0719$) and saved 2.083 questions (95\% CI 2.044 to 2.125) relative to fixed six items (Figure~\ref{fig:tradeoff}).

\subsection{Symptom-level errors}

Joint scoring was most accurate for stress (QWK 0.977), mood swing (0.963), bloating (0.939), and cramps (0.928). Sleep issues reached 0.867. Fatigue was the principal residual error, with QWK 0.696, MAE 0.528, exact agreement 62.50\%, and moderate-or-higher recall 56.88\%. Most fatigue errors shifted zero to low severity or shifted moderate fatigue down to low severity. The sleep--energy question therefore accounted for most of the agreement loss in the three-cluster policy.

\begin{figure}[htbp]
  \centering
  \includegraphics[width=\textwidth]{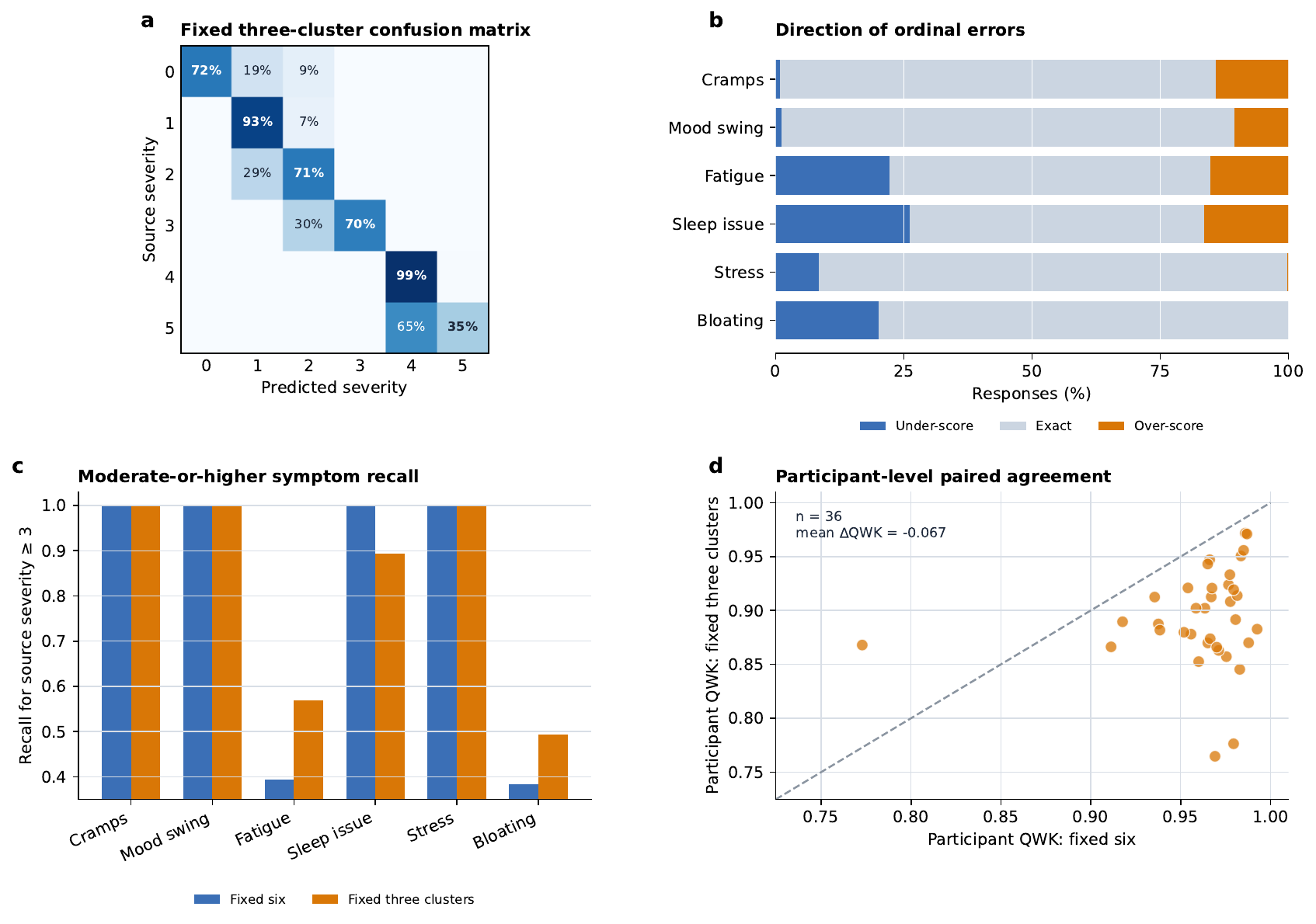}
  \caption{Error structure of fixed three-cluster questioning. \textbf{a}, row-normalized six-level confusion matrix. \textbf{b}, proportion of under-scored, exact, and over-scored labels by symptom. \textbf{c}, recall for source severity 3--5 under fixed six-item and fixed three-cluster policies. \textbf{d}, participant-level QWK under the two fixed policies; the dashed diagonal denotes equal agreement. Fatigue and bloating showed the lowest moderate-or-higher recall, while most participants remained close to the high-agreement region.}
  \label{fig:errors}
\end{figure}

\section{Discussion}

The main result is operational: three active symptom-cluster questions recovered all six daily labels with 97.45\% within-one agreement and cut the question count in half. Fixed six-item questioning remained the accuracy reference, but the cluster policy increased recall for moderate-or-higher symptoms. The compact policy therefore exchanges exact ordinal precision for lower repeated burden, and the bootstrap interval quantifies that exchange rather than hiding it behind a single accuracy value.

Open-first dialogue performed worse than direct cluster elicitation. An unrestricted opening exposed only a small fraction of the six target domains, forcing almost complete follow-up while also introducing evidence-gate errors. This finding supports a constrained conversational architecture: the system should ask natural topic-level questions, but the coverage controller and output schema should remain deterministic. The result extends topic-based questionnaire completion \citep{navarro2026questionnaire} by measuring label recovery and daily question burden on a locked participant split.

The comparison with conversational PHQ-9 work is also informative. HopeBot preserved a fixed questionnaire through interactive clarification and achieved high score concordance \citep{guo2026hopebot}; our best fixed six-item condition similarly shows that a small local scorer can preserve structured labels. The three-cluster condition tests a more aggressive compression mechanism: one answer fills two separately scored fields. Its strongest results occurred when the paired constructs were linguistically distinct, as with mood swing and stress. Fatigue and sleep were harder to separate because everyday descriptions of poor sleep, low energy, and functional slowing overlap.

The next experiment is a paired prospective study in which participants complete both the original daily ratings and the three-question dialogue across menstrual cycles. The model and prompts from this paper provide the frozen comparator. The only planned algorithmic change before that study is a revised sleep--energy exchange that first asks for sleep disruption and then distinguishes next-day tiredness from independent fatigue when the response is ambiguous.

\section{Conclusion}

Daily premenstrual symptom labels can be collected through a controlled three-question dialogue using small local language models. The fixed cluster policy reduced questions by 50\%, retained complete label coverage, and achieved QWK 0.913. Active structured elicitation was more efficient and more reliable than beginning with an unconstrained open prompt.

\section*{Data and code availability}

mcPHASES is available from PhysioNet under its data use agreement \citep{lin2025physionet}. Experiment configuration, prompts, evaluation code, figure source data, and frozen run manifests are included in the accompanying repository. An anonymized repository URL will be inserted before public submission.

\section*{Acknowledgments}

To be completed.

\section*{Funding}

To be completed.

\section*{Competing interests}

The authors declare no competing interests.

\appendix

\section{Symptom clusters and elicitation targets}

All model inputs and generated utterances were in Simplified Chinese. The three fixed clusters corresponded to the following English translations: ``How have your mood and sense of pressure been over the past two days? Have you been more irritable, tense, or less tolerant of small problems than usual?''; ``How have you been sleeping, and can your energy carry you through the day, or do ordinary tasks already feel tiring?''; and ``Have you had abdominal discomfort, such as heaviness, cramping, swelling, or clothes feeling tighter?'' The scorer returned two independent integer labels for each response.

\section{Reproducibility summary}

The development participant IDs, locked participant IDs, random seeds, model identifiers, severity anchors, prompt version, and dialogue-policy version are stored in the run configuration. Final locked evaluation sampled 10 days for each of 36 participants with seed 20260812. Participant bootstrap inference used 2,000 replicates with seed 20260813. Generation used seed 20260810, and evidence-gate training used seeds 20260809--20260811.


\begin{thebibliography}{99}
\raggedright
\setlength{\itemsep}{0.35em}

\bibitem[Endicott et~al.(2006)Endicott, Nee, and Harrison]{endicott2006drsp}
Endicott J, Nee J, Harrison W.
Daily Record of Severity of Problems (DRSP): reliability and validity.
\emph{Archives of Women's Mental Health}. 2006;9(1):41--49.
\doi{10.1007/s00737-005-0103-y}

\bibitem[Eisenlohr-Moul et~al.(2017)Eisenlohr-Moul, Girdler, Schmalenberger, Dawson, Surana, Johnson, and Rubinow]{eisenlohr2017cpass}
Eisenlohr-Moul TA, Girdler SS, Schmalenberger KM, et al.
Toward the reliable diagnosis of DSM-5 premenstrual dysphoric disorder: the Carolina Premenstrual Assessment Scoring System (C-PASS).
\emph{American Journal of Psychiatry}. 2017;174(1):51--59.
\doi{10.1176/appi.ajp.2016.15121510}

\bibitem[Bosman et~al.(2016)Bosman, Jung, Miloserdov, Schoevers, and aan het Rot]{bosman2016daily}
Bosman RC, Jung SE, Miloserdov K, Schoevers RA, aan het Rot M.
Daily symptom ratings for studying premenstrual dysphoric disorder: a review.
\emph{Journal of Affective Disorders}. 2016;189:43--53.
\doi{10.1016/j.jad.2015.08.063}

\bibitem[Epstein et~al.(2017)Epstein, Lee, Kang, Agapie, Schroeder, Pina, Fogarty, Kientz, and Munson]{epstein2017tracking}
Epstein DA, Lee NB, Kang JH, et al.
Examining menstrual tracking to inform the design of personal informatics tools.
In: \emph{Proceedings of the 2017 CHI Conference on Human Factors in Computing Systems}; 2017:6876--6888.
\doi{10.1145/3025453.3025635}

\bibitem[Harrison et~al.(2023)Harrison, Trickett, Wormald, Dobbs, Lis, Popov, Beard, and Rodrigues]{harrison2023emcat}
Harrison C, Trickett R, Wormald J, et al.
Remote symptom monitoring with ecological momentary computerized adaptive testing.
\emph{Journal of Medical Internet Research}. 2023;25:e47179.
\doi{10.2196/47179}

\bibitem[Lin et~al.(2026)Lin, Li, Kalani, Truong, and Mariakakis]{lin2026mcphases}
Lin G, Li JY, Kalani K, Truong KN, Mariakakis A.
A longitudinal dataset of physiological, hormonal, metabolic, and self-reported menstrual health data.
\emph{Scientific Data}. 2026.
\doi{10.1038/s41597-026-06805-3}

\bibitem[Lin et~al.(2025)Lin, Li, Kalani, Truong, and Mariakakis]{lin2025physionet}
Lin B, Li JY, Kalani K, Truong K, Mariakakis A.
mcPHASES: a dataset of physiological, hormonal, and self-reported events and symptoms for menstrual health tracking with wearables.
PhysioNet, version 1.0.0; 2025.
\doi{10.13026/zx6a-2c81}

\bibitem[Sheng et~al.(2023)Sheng, Finzel, Lucke, Dufresne, Gini, and Pakhomov]{sheng2023dialogue}
Sheng Z, Finzel R, Lucke M, Dufresne S, Gini M, Pakhomov S.
A dialogue system for assessing activities of daily living: improving consistency with grounded knowledge.
In: \emph{Proceedings of the Third DialDoc Workshop}; 2023:68--79.
\doi{10.18653/v1/2023.dialdoc-1.8}
\endgraf\smallskip

\bibitem[Liu et~al.(2025)Liu, Gu, Tong, Yue, Qiu, Zeng, Yu, Yang, and Zhao]{liu2025agreement}
Liu J, Gu J, Tong M, et al.
Evaluating the agreement between ChatGPT-4 and validated questionnaires in screening for anxiety and depression in college students.
\emph{BMC Psychiatry}. 2025;25:359.
\doi{10.1186/s12888-025-06798-0}

\bibitem[Guo et~al.(2026)Guo, Lai, Ive, Petcu, Wang, Qi, Thygesen, and Li]{guo2026hopebot}
Guo Z, Lai A, Ive J, et al.
Feasibility and user evaluation of HopeBot: an LLM-powered conversational chatbot for depression screening.
\emph{PLOS Digital Health}. 2026;5(6):e0001446.
\doi{10.1371/journal.pdig.0001446}

\bibitem[Fraile Navarro and Peleg(2026)]{navarro2026questionnaire}
Fraile Navarro D, Peleg M.
Conversational AI for automated patient questionnaire completion: development insights and design principles.
\emph{arXiv preprint arXiv:2602.19507}. 2026.
\doi{10.48550/arXiv.2602.19507}

\bibitem[Qwen Team(2025)]{qwen2025technical}
Qwen Team.
Qwen2.5 technical report.
\emph{arXiv preprint arXiv:2412.15115}. 2025.
\doi{10.48550/arXiv.2412.15115}

\bibitem[Warner et~al.(2024)Warner, Chaffin, Clavi\'e, Weller, Hallstr\"om, Taghadouini, Gallagher, Biswas, Ladhak, Aarsen, Cooper, Adams, Howard, and Poli]{warner2024modernbert}
Warner B, Chaffin A, Clavi\'e B, et al.
Smarter, better, faster, longer: a modern bidirectional encoder for fast, memory efficient, and long context finetuning and inference.
\emph{arXiv preprint arXiv:2412.13663}. 2024.
\doi{10.48550/arXiv.2412.13663}

\bibitem[Cohen(1968)]{cohen1968weighted}
Cohen J.
Weighted kappa: nominal scale agreement with provision for scaled disagreement or partial credit.
\emph{Psychological Bulletin}. 1968;70(4):213--220.
\doi{10.1037/h0026256}

\bibitem[Efron and Tibshirani(1993)]{efron1993bootstrap}
Efron B, Tibshirani RJ.
\emph{An Introduction to the Bootstrap}.
New York: Chapman and Hall/CRC; 1993.
\doi{10.1201/9780429246593}

\end{thebibliography}
\end{document}